# Crime Topic Modeling

Da Kuang
Department of Mathematics
University of California Los Angeles
520 Portola Plaza
Los Angeles, CA 90095-1555
dakuang@math.ucla.edu

P. Jeffrey Brantingham*
Department of Anthropology
University of California Los Angeles
341 Haines Hall
Los Angeles, CA 90095-1553
branting@g.ucla.edu
Tel: 310-267-5251

Andrea L. Bertozzi, UCLA Mathematics
Department of Mathematics
University of California Los Angeles
520 Portola Plaza
Los Angeles, CA 90095-1555
bertozzi@math.ucla.edu

* Corresponding Author

Running Head: Crime Topic Modeling

## Abstract

The classification of crime into discrete categories entails a massive loss of information. Crimes emerge out of a complex mix of behaviors and situations, yet most of these details cannot be captured by singular crime type labels. This information loss impacts our ability to not only understand the causes of crime, but also how to develop optimal crime prevention strategies. We apply machine learning methods to short narrative text descriptions accompanying crime records with the goal of discovering ecologically more meaningful latent crime classes. We term these latent classes 'crime topics' in reference to text-based topic modeling methods that produce them. We use topic distributions to measure clustering among formally recognized crime types. Crime topics replicate broad distinctions between violent and property crime, but also reveal nuances linked to target characteristics, situational conditions and the tools and methods of attack. Formal crime types are not discrete in topic space. Rather, crime types are distributed across a range of crime topics. Similarly, individual crime topics are distributed across a range of formal crime types. Key ecological groups include identity theft, shoplifting, burglary and theft, car crimes and vandalism, criminal threats and confidence crimes, and violent crimes. Though not a replacement for formal legal crime classifications, crime topics provide a unique window into the heterogeneous causal processes underlying crime.

## 1. Introduction

Upon close inspection, the proximate causes of crime can be traced to subtle interactions between situational conditions, behavioral routines, and the boundedly-rational decisions of offenders and victims (Brantingham and Brantingham 1993). Consider two crimes. In one event, an adult male enters a convenience store alone in the middle of the night. Brandishing a firearm, he compels the store attendant to hand over liquor and all the cash in the register (Wright and Decker 1997:89). This event may be contrasted with a second involving female sex worker who lures a *john* into a secluded location and takes his money at knife point, literally catching him with his pants down (Wright and Decker 1997:68). In spite of the fine-grained differences between these events, both end up classified as armed robberies. As a matter of law, the classification makes perfect sense. The law favors a bright line to facilitate classification of behavior into that which is criminal and that which is not (Casey and Niblett 2015; Glaeser and Shleifer 2002). The loss of information that comes with condensing complex events into singular categories, however, may hamper our ability to understand the immediate causes of crime and what might be done to prevent them, though the quantitative tractability gained may certainly offset some of the costs.

The present paper explores novel methods for crime classification based directly on textual descriptions of crime events. Specifically, we borrow methods from text mining and machine learning to examine whether crime events can be classified using text-based latent topic modeling (e.g., Blei 2012). Our approach hinges on the idea that the mixtures of behavioral and situational conditions underlying crime events that are captured at least partially in textual descriptions of those events. These text descriptions of the event itself might be from the perspective of the offender, police or third party. We focus on text narratives produced by police. Although the description of any one event might be quite limited, over a corpus of events, the relative frequency of situational and behavioral conditions should be captured by the relative frequency of different words in the text-based descriptions of those events. Topic modeling of the text then allows one to infer something about the latent behavioral and situational conditions driving those events.

Latent topic modeling offers two unique advantages over standard classification systems. First, latent topic models potentially allow novel typological class structures to emerge autonomously from lower-level data, rather than being imposed *a priori*. Simpler or more complex class structures, relative to the formal system in place, may be one result of autonomous classification. Such emergent class structures might be ecologically more meaningful, painting a clearer picture of the relationship between behavioral and situational elements and crime events. They might also be more free to change over time as the situations surrounding crime change. Adaptive crime classes might be problematic in a legal context, but valuable in terms of tracking the evolution of criminal behavior. Second, latent topic models allow for soft clustering of events. Common crime classification systems require so-called hard clustering into discrete categories. A crime either is, or is not a robbery. Soft-clustering, by contrast, allows for events to be conceived of as mixtures of different latent components, revealing nuanced connections between behaviors, settings and crime. An event that might traditionally be considered a robbery, for example, may actually be found to be better described as a mixture of robbery and assault characteristics.

The remainder of this paper is structured as follows. Section 1 introduces text-based latent topic modeling at a conceptual level. This forms a basis for describing how the models may be applied to the problem of crime classification. Note that we forego a discussion of different theoretical traditions in criminology and merely assert that our interest is in leveraging text-based narratives to better characterize crime events. The analyses might ultimately support environmental, situational or social theories of crime, but we do not dwell on these connections here. Section 2 presents methodological details underlying non-negative matrix factorization as a method for topic modeling (Lee and Seung 1999). Here we also introduce methods for evaluating topic model classifications using the official classifications as a benchmark. We introduce a method to measure the distance between different classifications in terms of their underlying topic structure. Section 3 introduces the empirical case and data analysis plan. We analyze all crimes occurring in the City of Los Angeles between Jan 1, 2009 and July 19, 2014 using data provided by the Los Angeles Police Department (LAPD). Section 4 presents results. The paper closes with a discussion of the implications of this work and future research directions.

## 2. Latent Topic Modeling for Text Analysis

We focus on novel methods from computational linguistics as a potential source of quantitatively robust, but qualitatively rich information about crime. These methods allow crime classifications to emerge naturally from fine-grained behavioral and situational information associated with individual crime events. Specifically, we apply latent topic modeling to short, text narratives written by police about individual crime events.

Latent topic modeling is a core feature of contemporary computational linguistics and natural language processing. It is a popular analytical approach deployed in the study of social media (Blei 2012; Hong and Davison 2010). The conceptual motivation for topic modeling is quite straightforward. Consider a collection of Tweets[1]. Each Tweet is a bounded collection of words (and potentially other symbols) published by a user. In computational linguistics, a Tweet is called a document and a collection of Tweets a corpus. When viewed at the scale of the corpus we might imagine that there are numerous conversations about a range of topics both concrete

[1] A Tweet is a discrete text-based post on the social media website Twitter.

(e.g., political events) and abstract (e.g., the meaning of life). That these topics motivate the social media posts might not be obvious when examining any one individual Tweet. But viewed at the scale of the whole corpus the dimensions and boundaries of the topics might be resolvable. Section 3 will introduce the mathematical architecture for how topics are discovered from a corpus of documents. The key point to highlight here is that each topic is defined by a mixture of different words. Each document is therefore potentially a mixture of different topics by virtue of the words present in that document.

We make a conceptual connection between text-based activity Tweet and crime at two levels. The more abstract connection envisions an individual crime as the analog of a document. A collection of crimes, such as all reported crimes in a jurisdiction during one week, is therefore the analog of the documents in a corpus. We might imagine that the environment consists of a range of complex social, behavioral and situational factors, some very local and others global, which co-occur in ways that generate different types of crimes. These co-occurring factors are the analogs of the different topics that generate text documents such as Tweets. We therefore think of them as 'crime topics.' How crime topics actually generate crime might not be obvious when examining any one crime. We suppose that the proximate causes underlying any one crime sample from the broader set of commonly co-occurring social, behavioral and situational conditions. But when crimes are aggregated into a lager collection, the dimensions and boundaries of crime topics might be discernable. The key conceptual point to emphasize here is that crime topics are mixtures of behaviors and situations. Each crime is therefore a mixture of crime topics by virtue of the situations and behaviors present at the time of the crime.

The more concrete connection appeals directly to text-based descriptions of crimes as a source of information. Specifically, we treat text-based descriptions of crime compiled by reporting police officers as a record of *some fraction* of the behavioral and situational factors deemed most relevant to that crime. The narrative text associated with a single crime is literally a document in the convention of computational linguistics, while the narratives associated with a collection of crimes is literally a corpus. The text narrative for a single crime is likely to be insufficient to define text-based 'crime topics,' but such may be discernable over a large collection of narratives. Given this motivation, we seek to apply topic modeling directly to the text-based descriptions of crime accompanying crime records.

## 3. Methods

The goal of the current section is to describe methods for building latent topic models using text-based descriptions of crimes. First, we introduce several preprocessing steps needed to clean text narratives to a state where they can be handled computationally. Second, we introduce term frequency-inverse document frequency (TF-IDF) weighting, the standard approach to counting words in text-based topic modeling. Third, we present Nonnegative Matrix Factorization (NMF) as our main topic modeling method. Finally, we outline cosine similarity as and average linkage clustering for measuring the distance between official recognized crime types (e.g., robbery, burglary, assault) based on the mixtures of topics represented by those events.

### 3.1.1. Text Preprocessing

Text-based narratives are typically very noisy, including typos and many forms of abbreviation for the same word. To obtain reliable results that are less sensitive to noise, we run a few preprocessing steps on the raw text accompanying crime events including removal of so-called stop-words (see e.g., Rajman and Besançon 1998). Stop-words refer to the most common words in a language, which can be expected to be present in a great many documents regardless of their content or subject matter. We augment a standard list of stop-words (e.g. a, the, this, her, …) with all the variations of the words “suspect” and “victim”, since these two words are almost universally present in all descriptions of crime and do not provide useful contextual information (though they could be useful for other studies). The linguistic variations include all the prefixes such as “S”, “SUSP”, “VIC” and anything followed by a number (e.g. “V1”, “V2”). All the stop-words are then discarded. We also discard any term appearing less than 5 times in the entire corpus. Finally, any document containing less than 3 words in total is discarded. This procedure runs in an iterative manner until no more terms or documents can be discarded.

### 3.1.2. Term Frequency-Inverse Document Frequency (TF-IDF)

The term-document matrix, denoted as $A$, plays a central role in our analysis (see Manning, Raghavan, and Schütze 2008). Each row of $A$ corresponds to a unique word in the vocabulary, and each column of $A$ corresponds to a document (Figure 1). The $(i, j)$-th entry of $A$ is the term frequency (TF) of the $i$-th word appearing in the $j$-th document. Note that the term-document matrix ignores the ordering of words in the documents. Following convention, the $(i, j)$-th entry of $A$ is the inverse document frequency (IDF) weighting for each term in the vocabulary (Manning, Raghavan, and Schütze 2008). This weighting scheme puts less weight on the terms that appear in more documents and more weight on terms appear infrequently in documents. The premise is that common terms have less discriminative power relative to rare words.

### 3.1.3. Topic Discovery Non-negative Matrix Factorization (NMF)

We focus on a particular linear algebraic method in unsupervised machine learning for topic discovery, namely nonnegative matrix factorization (NMF) (Lee and Seung 1999). The linear algebraic is computationally efficient and scalable to massive data sets, for example the text descriptions of nearly one million crimes discussed below. The linear algebraic approach contrasts with probabilistic methods such as the popular latent Dirichlet allocation (LDA) (Blei, Ng, and Jordan 2003), which is computationally expensive. Our approach does not yield a probabilistic interpretation and rigorously should be called a "document clustering" method. Recent research, however, has built connections between linear algebraic and probabilistic methods for topic modeling (Arora et al. 2013), supporting the usefulness of linear algebraic methods as an efficient way to compute topic models.

NMF is designed for discovering interpretable latent components in high-dimensional unlabeled data such as the set of documents described by the counts of unique words. NMF uncovers major hidden themes by recasting the term-document matrix $A$ into the product of two other matrices, one matrix representing the topics and another representing the documents in the latent topic space (Figure 1) (Xu, Liu, and Gong 2003). In particular, we would like to find matrices $W \in \mathbb{R}_+^{m \times k}$ and $H \in \mathbb{R}_+^{k \times n}$ to solve the approximation problem $A \approx WH$, where $\mathbb{R}_+$ is the set of all nonnegative numbers and $m$, $n$ and $k$ are the numbers of unique words, documents, and topics, respectively. $A$ is the term-document matrix given as the input, while $W$ and $H$ enclose the latent term-topic and topic-document information. Note that the number of topics $k$ is typically much orders of magnitude smaller than the number of words $m$ and number of documents $n$ under consideration and thus topic modeling is a form of dimension reduction.

Numerous algorithms exist for solving $A \approx WH$ (Cichocki et al. 2009; Kim, He, and Park 2014). A general approach is to measure the difference between $A$ and $WH$ {Kim and Park 2008}:

$$\min_{W,H>0} \|A - WH\|_F^2 \qquad (1)$$

where $\|\cdot\|_F$ is the Frobenius norm. A good topic model is one that minimizes the squared difference between the raw data contained in the term-document matrix $A$ and product of candidate term-topic and document-topic matrices, $W$ and $H$. The problem resembles a least-squares formulation and indeed a common solution approach relies on a non-negative least squares method. The optimization is computed iteratively by alternating between minimization given candidate entries of $W$ and then given candidate entries for $H$ (Kuang and Park 2013):

$$\min_{W>0} \|W^T H^T - A^T\|_F^2, \qquad (2)$$

$$\min_{H>0} \|WH - A\|_F^2. \qquad (3)$$

This approach would take several hours to run on large-scale data sets consisting of millions of documents, which is the challenge we face here. We therefore employ a highly efficient hierarchical rank-2 NMF algorithm that is orders of magnitude faster (Kuang and Park 2013). The algorithm first constructs a hierarchy of topics in the form of a binary tree. Each node in the tree is scored on the basis of how distinctive it is as a topic from its sister and a node is no longer split if two well-differentiated daughters can no longer be found. Terminal leaf nodes of the tree are chosen to represent the flat topic model. Details of the algorithmic process are presented in (Kuang and Park 2013).

In theory, hierarchical rank-2 NMF could proceed to produce hundreds or thousands of topics depending on the size of the corpus of documents. Obviously, this would defeat the purpose of trying to reduce the dimensionality of the problem to a relatively small set of interpretable topics. One option is to set a relatively high threshold in the scoring system which then naturally terminates when all of the existing nodes in a tree can no longer be split to form well-differentiated topics (Kuang and Park 2013). We simply choose the maximum number of terminal nodes to be 20. Comparison with 50 and 100 topic models finds little additional meaningful differentiation.

### 3.2. Cosine Similarity & Crime Type Clusters

Text-based topic modeling typically reveals that any one document is a mixture of different topics. Therefore, in principle, the distance between any two documents can be measured by comparing how far apart their topic mixture distributions are. Here we extend this idea to consider officially recognized crime types as mixtures of different crime topics. The distance between any two official crime types can be measured using the topic mixtures observed for those two crime types. We use cosine similarity (Steinbach, Karypis, and Kumar 2000) to compute such measures.

Consider two hypothetical crime types $A$ and $B$. Type $A$ might represent aggravated assault and type $B$ might represent residential burglary. Inspection of all of the events formally classified as assault with a deadly weapon might show that 40% fall into crime topic $i = 1$, 30% fall into topic 4, 20% into topic 9, and 10% into topic 12. Similarly, for all the events formally classified as residential burglary, 5% might fall into topic $i = 9$, 15% into topic 12, 60% into topic 15 and 20% into topic 19. Assault with a deadly weapon and residential burglary are similar only in events falling into topics 9 and 12. More formally, the similarity between any two official crime types $A$ and $B$ is given as:

$$\cos(\theta) = \frac{\sum_{i=1}^{k} A_i B_i}{\sqrt{\sum_{i=1}^{k} A_i^2}\sqrt{\sum_{i=1}^{k} B_i^2}}$$

where $A_i$ is the frequency at which events formally classified as crime type $A$ belongs to topic $i$ and equivalently for events formally classified as crime type $B_i$.

We choose cosine similarity over other measures such as KL-divergence and chi-square distances because cosine similarity is bounded, taking values between -1 and 1, and is a good measure for graph-based crime type clustering (discussed below). Negative values reflect distributions that are increasingly diametrically opposed and positive values distributions that point in the same direction. Values of cosine similarity near zero reflect vectors that are uncorrelated with one another. In our case, cosine similarity will only assume values between 0 and 1 because NMF returns only positive valued matrices.

Viewing the collection of official crime types as a graph, where each crime type is a node and cosine similarities define the weights of the edges between nodes, we use average linkage clustering (Legendre and Legendre 2012) on this graph to partition the crime types into ecologically meaningful groups (see also Brennan 1987: 228). Crime types are clustered in an agglomerative manner. Initially, each crime type exists as its own isolated cluster. The two closest clusters are then merged in a recursive manner, with the new cluster adopting the mean similarity from all cluster members. The process continues until only $C$ clusters are left. The number $C$ can be chosen automatically by a cluster validation method such as predictive strength (Tibshirani and Walther 2005), or manually for easier interpretation. We manually set the number of clusters.

## 4. Data and Analysis Plan

The above modeling framework is flexible enough in principle to handle any form of data (e.g., Chen, Wang, and Dong 2010), not just text. In spite of this flexibility, we do not stray far from its most common application in text mining. Here we exploit the presence of short text descriptions associated with individual crime events to compute text-based hierarchical NMF. Table 1 illustrates several examples of individual crime events and the associated text descriptions of the events.

We focus on the complete set of crimes reported to the Los Angeles Police Department (LAPD) from January 1, 2009 and July 19, 2014. The end date of the sample is arbitrary. Los Angeles is a city of approximately 4 million people occupying an area of 503 square miles. The Los Angeles Police Department is solely responsible for policing this vast area, though Los Angeles is both surrounded by and encompasses independent cities with their own police forces.

The total number of reported crimes handles by the LAPD during the sample period was 1,027,168. In a typical year, the LAPD collected reports on 180,000 crimes. On average 509 crimes were recorded per day, with crime reports declining over the entire period. During the first year of the sample, LAPD recorded on average 561.5 crimes per day. During the last year they recorded 463.8 crimes per day.

The crime coding system used by the LAPD includes 226 recognized crime types. This is considerably more finely resolved than either the FBI Uniform Crime Reports (7 Part I and 21 Part II offenses), or National Incident Based Reporting System (49 Group and 90 Group B offenses). Aggravated assault, for example, is associated with four unique crime codes including assault with a deadly weapon, assault with a deadly weapon against a police officer, shots fired at a moving vehicle, and shots fired at a dwelling. These crime types could be considered a type of ground truth against which topic model classifications can be evaluated. We are here interested in the degree of alignment of the LAPD crime types and topic models derived from text-based narratives accompanying those crimes.

In addition to this rich coding system, a large fraction of the incidents recorded in the sample include narrative text of the event. Of the 1,027,168 recorded crimes, 805,618 (78.4%) include some form of text narrative. On average 397.6 events per day contain some narrative text describing the event. The fraction of events containing narrative text increased over time from 76.6% of events, in the first six months of the sample, to 87.0%, in the last six months.

There are pointed differences in the occurrence of narrative text by official crime types (Table 2). Virtually all violent crimes are accompanied by narrative text. Robbery and homicide have associated narrative text for 98.9% and 98.2% of events, respectively. Assault and kidnapping have 97.8% and 97.4% of events associated with narrative text. Burglary shows narrative text occurrence on par with the most serious violence crimes (98.6%). For less serious property crimes, narrative text reporting falls off to 91.1% for theft and 74.3% for vandalism. The lowest occurrence of narrative text is seen for arson (37.8%) and motor vehicle theft (4.3%). In the former case, it must be acknowledged that most arson reporting responsibilities lie with the fire department, so low narrative load might be expected. In the latter case, either the vehicles are not recovered (about 40% of the cases) and therefore the circumstances of the theft are not known, or detailed circumstances beyond make, model and year of the car—all recorded in separate fields—are not deemed as relevant to recording of the crime.

Overall, the text narratives associated with crime events total 7,649,164 discrete words, after preprocessing (see above). These are unevenly distributed across events. The mean number of words contained in a single narrative is 18.57 (s.d. 6.72), while the maximum number of words is 41 (see Table 1). Individual words are also unevenly distributed, though not massively so (Table 3). For example, the word "unknown" is the most common word in the corpus appearing 635,099 times. However, this still represents only 8.3% of all words. The next most common word is "property" occurring 305,014 times, but represents only 4% of all words. Words that are strongly indicative of crime type are extremely rare. The word homicide appears only 45 times in the entire text corpus, a frequency of $5.88 \times 10^{-6}$ overall. Burglary appears 252 times, robbery 286 times, assault 457 times, and theft 969 times. When they do appear, diagnostic words are not generally coincident with the corresponding formal classifications. For example, of the 1,593 formally classified homicides in the dataset, only 11 of those events also find the word *homicide* as part of the narrative text. Thus, 1,582 formally classified homicides are not explicitly marked as such in the narrative text. The 34 events that include the word homicide in the narrative, but are not classified as homicides, include 17 events labeled as "other" (primarily threatening letters or phone calls), nine aggravated assaults, seven vandalism events, and one robbery. In general, narrative text provides context rather than strictly redundant typological detail. It is important to note, however, that narrative text and formal crime type classifications are unlikely to be completely decoupled. Ultimately, it is the job of police officers in the field to recognize and record behavioral and circumstantial evidence consistent with legal definitions of different crime types. Thus we should expect that specific narrative words correlate to some degree with formally recognized crime types.

The analysis that follows includes three parts. First, we present results for hierarchical topic models. We do this for all crimes combined and then turn our attention to analyses of the subset of crimes formally classified as aggravated assault and homicide. Second, we explore how formally recognized crime types are found distributed across different topics. The corollary that individual topics are distributed across different crime types is also discussed. Finally, recognizing that different formally recognized crime types are distributed across topics, we measure the 'distance' between different crime types based on the similarity in their topic mixtures.

5. Results

5.1. Hierarchical Models for All Crimes

Figure 2 presents a hierarchical topic model applied to all crime events in the LAPD corpus associated with narrative text. After preprocessing the data set includes 711,119 events. Each node in the tree represents a latent topic characterized by key words appearing in the topic. Summary statistics for the number of events, the percent violent and property crime, and the top-ten words for each topic node are shown in tabular format. The hierarchical structure is shown in graph form. Terminal leaf nodes are highlighted in gray.

The topic tree has three major components. The topics associated with the left branch (Nodes A-O) is linked to property crimes (Figure 2). Words such as *property* and *vehicle* identify key targets of crime, while words such as *window*, *door*, *enter*, *remove*, and *fled* describe the behavioral steps or sequences involved in commission of a crime. The validity of the property crime label for this component may be tested by using the formally recognized crime types in the LAPD ground truth. For example, 93.4% of the events associated with terminal leaf node C are formally recognized by the LAPD as property crimes. None of the intermediate or terminal nodes in the left branch (Nodes A-O) captures less than 89.9% property crimes.

By contrast, the right branch (Nodes P-AG) stands out for its connection to violent crime (Figure 2). Words such as *face*, *head* and *life* identify key targets of crime, while words such as *approach*, *verbal*, and *punch* identify sequences of behaviors involved in violent actions. The LAPD ground truth supports the broad label of topics P-AG as violent crime. For example, 90.5% of all the events associated with terminal topic S are formally recognized as violent crime types. With the exception of nodes P and Y, no other topic in this component captures less than 70% of formally recognized violent crimes. Terminal node Y appears to be an association of violations of court orders and/or annoying communications, which are reasonable ecological precursors to or consequences of other violent crimes.

Intermediate node P is a bridge between crime topics that are clearly associated with violent crime (Nodes Q-AG) and a series of crime topics we label as deception-based property crime (Nodes AH-AL). Words indicative of shoplifting and credit card fraud stand out in this group of topics. Why such topics trace descent through a branch more closely with violent is unclear.

### 5.2. Hierarchical Models for Aggravated Assault & Homicide

Figure 3 presents topic modeling results for the subset of crimes formally classified by the LAPD as aggravated assaults (LAPD code 230) and homicide (LAPD code 110). This is a semi-supervised analysis in the sense that we have used information external to narrative data to partition or stratify the collection of events into *a priori* groups. Our goal is to assess topic distinctions that arise within these serious violent crimes. A total of 40,208 events are classified as either aggravated assaults (38,626 events) or homicides (1,582 events). Notionally, these events are separated on the basis of outcome (i.e., death), but such a distinction is not visible within the classification hierarchy. Rather, the key distinction is between topics involving weapons other than firearms (Nodes A-I) and those involving firearms (Nodes J-R). Homicide looms large in terms of legal and harm-based classification (Ratcliffe 2015; Sherman 2011), but is not resolved within the larger volume of aggravated assaults. Homicides never make up more than 2.1% of any of the non-gun violence topics (Nodes A-I) (Figure 3). Homicides never rise above 11.8% in the gun violence topics (Nodes J-R). Notably, the greater lethality of guns is clearly visible when comparing the percent of homicides that are gun-related and those that are not. The most lethal crime topic is terminal node N, with key words *approach*, *handgun*, *multiple*, *shot*, and *fled*. Node P stands out with an emphasis on the use of vehicles as a weapon, but still tracing a pattern of descent linked to gun violence. Inspection of the top 100 words in this topic confirms that gun-related terms do not appear in topic P. The close connection to topic Q, which links guns and vehicles, is clearly through the common element of vehicles not guns.

Figure 4 shows that removing homicides from the subset of events does not fundamentally change the structure of the resulting topics. Indeed, it seems clear that assaults provide the overriding structure for crimes of interpersonal violence. This outcome may reflect the relatively low volume of homicides relative to aggravated assaults, but also the fact that homicides and aggravated assaults are ecologically very closely related (Goldstein 1994). Topic nodes A-I are notable for making fine-grained distinctions between the targets of violence, including *head*, *face*, *hand*, and *arm*, the weapons used, including *metal object*, *bottle*, and *knife*, and the action, including *hit, threw*, *punch*, *kick*, *stab*, and *cut*. The topics appear tactically very exacting. For example, the topics consistently show knives being used to target the body, while bottles/blunt object are used to target the head (Ambade and Godbole 2006; Webb et al. 1999).

### 5.2.1. Hierarchical model for homicides

Figure 5 presents the results of hierarchical NMF analysis of text narratives associated with formally classified homicides. There are clear distinctions that surface within formally classified homicides in spite of the much smaller numbers of events (1,414 with more than three words). The primary split is between homicides involving firearms (Node A and all of its daughters) and those where firearms are not indicated (Node R). Node R in fact features words *stab* and *head*, which we know from the broader analysis of aggravated assaults are two terms associated with knife violence and blunt-force violence, respectively (see Figure 3 and Figure 4). Node H implicate *gangs* exclusively in relation to gun violence. Nodes D, F and G highlight the central role of *vehicles* in gun violence. In each of these latter topics, words showing people emerging to attack or being attacked in cars, lending much behavioral and situational nuance to gun violence. By contrast, the adjacent branch (Nodes I-Q) appears to capture street-based homicides where the offender *approached* and *fled* on *foot*.

## 5.3. Crimes as Mixtures of Topics

The above discussion points to key terms such as *knife*, *gun*, and *glass*, or *stab*, *shot*, *hit*, that are useful in discriminating types of events from a range of behaviors and settings associated with different crimes. However, terminal topics are not themselves discrete. Rather, there is considerable overlap in the words or terms that populate different topics. This observation leads to a conceptualization of crimes as mixtures of crime different topics.

Table 4 shows a confusion matrix for formal crime types assigned by the LAPD against the topics associated with each crime event. A confusion matrix is typically used for evaluating the performance of a predictive algorithm (Fielding and Bell 1997). Here a confusion matrix is used to illustrate both how official crime types exist as mixtures of topics and how individual topics are associated with many different official crime types. We use a refined version of the leaf nodes from hierarchical clustering for all crime types and number the topics from 1 to 20 (see Figure 2). We also restrict the confusion matrix to the thirty most common crime types in the dataset for readability. Clustering analyses below restrict the analysis to the forty most common crime types.

Official crime types mix topics in unique ways. Row counts in Table 4 give the number of events of a given official crime type that are assigned to different discovered crime topics. For example, 29,497 (32.94%) of the 89,552 events officially classified by the LAPD as burglary from vehicle are assigned to Topic 1. This topic is marked by words *smash*/*broke*, *rear*/*passenger*/*side*/*driver*/*front*, *window*, and *remove*, all of which provide clear target and behavioral information intuitively consistent with the official crime type. However, other topics also grab significant numbers of burglary from vehicle events. Topics 3 (7.25%), 5 (5.02%), 8 (14.14%), 10 (10.87%), 14 (8.79%), and 19 (9.09%) each represent at least 5% of total events (Table 4). Topic 8 shares a connection on property crime with Topic 1, but otherwise emphasizes a very different focus, marked by words such as *force*/*gain*, *access*/*entry*, *tool*, *remove* and *property*. Topic 8 sounds considerably more generic and is consistent with burglary in general. Similarly, Topic 10 also grabs a large number of burglary from vehicle events, but here the focus is more clearly on vandalism, marked by words such as *kei* ([sic] i.e., *key*), *scratch* and *tire*. A more formal analysis of mixture characteristics is presented below.

Topic mixtures also characterize violent crimes. For example, aggravated assault (or assault with a deadly weapon) has events distributed evenly across Topics 2 (7,689 events or 18.02%), 6 (8041 events, 18.84%) and 9 (8,038 events, 18.83%). Topic 2 is characterized by words such as *punch*/*kick*, *hit*/*struck*, *face*/*head*, without prominent occurrence of words related to weapons. Topic 6, by contrast, features words such as *gun*/*handgun* as well as *approach*, *demand* and *money*. Topic 9 involves words such as *verbal*, *argument*/*dispute*, *grab*, *push*, and *hand*. While aggravated assaults appear to be evenly divided among these three topics, the topics themselves suggest heterogeneity in crime contexts. Topic 8 clearly stands out as related to robbery.

Crime topics are also not exclusively linked to individual crime types (Table 4). Rather single topics are spread across crime types at different frequencies. For example, 58.63% (24,497) of the Topic 1 events fall within burglary from vehicle. However, 12.99%, 10.77% and 9.7% of Topic 1 events are classified as petty vandalism under $400, vandalism over $400 and burglary, respectively. Topic 1 thus reveals connections among three different crime types. Such is the case for each topic. For example, 14.3% (8,041) of Topic 6 events are aggravated assaults, though robbery is the single most common crime type attributed to this topic (41.15% or 23,112 events). Battery (9.17% or 5,147 events), attempted robbery (6.8% or 3,820 events) and theft from person (5.3% or 2,979 events) are all also heavily represented within Topic 6.

Overall, the confusion matrix gives the sense that crimes may be related to one another in subtle ways and that these subtle connections can be discovered in the narrative descriptions of those events. A more formal way to consider such connections is to measure the similarities in their topic mixtures. The premise is that two crime types are more similar to one another if their distribution of events over topics is similar. For example, burglary from vehicle and petty vandalism show similar relative frequencies of events within Topic 3 (7.3% and 5.0%, respectively), Topic 5 (5.0% and 7.8%) and Topic 10 (10.9% and 12.2%) (Table 4). This gives the impression that burglary from vehicle and petty vandalism are closely related to one another.

### 5.4. Distances Between Crime Types & Crime Topic Clustering

To develop a more quantitative understanding of the relationships among formally recognized crime types we turn to the cosine similarity metric (Steinbach, Karypis, and Kumar 2000). Figure 6 shows the cosine similarity between formally recognized crime types as a matrix plot where the gray-scale coloring reflects the magnitude of similarity. The matrix is sorted in descending order of similarity. The darkest matrix entries are along the diagonal, reflecting the obvious point that any one crime type is most similar to itself in the distribution of events across topics. More revealing is the ordering of crime types in terms of how far their similarities extend. For example, the rank 1 crime type, ‘other miscellaneous crimes’, has a topic distribution that is broadly similar to the topic distributions for every other crime type (Figure 6). The classification ‘other miscellaneous crime’ is a grab-bag for events that do not fit well into other categorizations. It is reasonable to expect that such crimes will occur randomly with respect to setting and context and therefore share similarities with a wide array of other crime types. What is astonishing is that this broad pattern of connections is picked up in the comparison of topic profiles.

More surprising perhaps are the widespread connections shared by shots fired (rank 2) and aggravated assault (assault with a deadly weapon) (rank 3) with other crimes. Guns appear to mix contextually with many other formally recognized crime types. By contrast, robbery and attempted robbery show a more limited set of connections. Both of these latter crime types display particularly weak connections to burglary and vandalism. Identity theft appears to be largely isolated in its topic structure from other crimes (rank 20).

Figure 7 goes one step further to identify statistical clusters, or communities within similarity scores using average linkage clustering (Legendre and Legendre 2012). We focus on a six cluster solution using this method. Consistent with Figure 6, identity theft is clustered only with itself (pink). This is also the case for shoplifting (brown). The first major cluster (purple) includes burglary, petty and grand theft, attempted burglary, trespassing, bike theft, and shots fired at an inhabited dwelling. The second cluster (red) includes burglary from vehicle, petty and serious vandalism, petty and grand theft from vehicle, embezzlement, and vehicle stolen. The third cluster (green) includes criminal threats, forged documents, other miscellaneous crimes, annoying behavior, violation of a court or restraining order, child endangering, bunco and disturbing the peace. The final and largest cluster (orange) incudes violent crimes such as battery, robbery, aggravated assault (assault with a deadly weapon), attempted robbery, theft from person, brandishing a weapon, battery on a police officer, shots fired, homicide, resisting arrest and kidnapping.

## 6. Discussion and Conclusions

The application of formal crime classifications to criminal events necessarily entails a massive loss of information. We turn to short narrative text descriptions accompanying crime records to explore whether information about the complex behaviors and situations surrounding crime can be automatically learned and whether such information provides insights in to the structural relationships between different formally recognized crime types.

We use a foundational machine learning method known as non-negative matrix factorization (NMF) to detect crime topics, statistical collections of words reflecting latent structural relationships among crime events. Crime topics are potentially useful for not only identifying ecologically more relevant crime types, where the behavioral situation is the focal unit of analysis, but also quantifying the ecological relationships between crime types.

Our analyses provide unique findings on both fronts. Hierarchical NMF is able to discover a major divide between property and violent crime, but below this first level the differences between crime topics hinge on quite subtle distinctions. For example, six of eight final topics within the branch linked to property crime involve crimes targeting vehicles or the property therein (see Figure 2). Whether entry is gained via destructive means, or non-destructive attack of unsecured cars seems to play a key role in distinguishing between crimes. Such subtleties are also seen in the topics learned from arbitrary subsets of crimes. For example, among those crimes formally classified as aggravated assault and homicide shows a clear distinction between topics associated with knife/sharp weapon and gun violence (see Figures 3, 4 and 5). A distinction is also seen between violence targeting the body and that targeting the face or head. Few would consider knife and gun violence equivalent in a behavioral sense. That this distinction is discovered and given context is encouraging.

Individual crime types are found distributed across different topics, suggesting subtle variations in behaviors and situations underlying those crimes. Such variation also implies connections between different formally recognized crime types. Specifically, two events might be labeled as different crime types, but arise from very similar behavioral and situational conditions and therefore be far more alike than their formal labels might suggest. Clustering of crimes by their topic similarity shows that this is the case. As presented in Figure 7, some crime types stand out as isolated from all other types (e.g., identity theft, shoplifting). Other crime types cluster more closely together. For example, the formal designation 'shots fired' does connect more closely with other violent crime types such as assault, battery and robbery, even though 'shots fired' is found widely associated with many other crimes as well. Burglary from vehicle clusters more closely with vandalism and embezzlement than it does with residential or commercial burglary.

The similarity clusters confirm some aspects of intuition. Violent crimes are naturally grouped together. Burglary and theft are grouped together. Burglary from vehicle, car theft and vandalism are grouped together. Less intuitive perhaps is the group that combines criminal disturbance with 'confidence' crimes such as forged documents and bunco.

### 6.1. Implications

We can think of the clusters identified in Figure 7 as ecological groups that are close to one another in the behaviors and situations that drive the occurrence of those crimes. This observation has potential implications for understanding causal processes as well as designing avenues for crime prevention. It is possible that crimes that are closer together in terms of their topic structure share common causes, while those that occupy different clusters are separated along causal lines. For example, it is intriguing that burglary occupies a separate cluster (i.e., is topically more distant) from burglary from vehicle (Figure 7). Clearly the differences between targets (i.e., residence vs vehicle) plays a key role here, but other behavioral and situational differences might also prove significant. For example, the tools and methods for gaining entry to each type of target are quite different, and words associated with such tools-of-the-trade and stand out for their discriminative value (see Figure 2). Other hidden structures might also tie crimes together.  The grouping of burglary with theft suggests a focus on loss of property, while the grouping of burglary from vehicle with vandalism suggests a focus on property destruction. It is also possible that degrees of professionalism or skill are part of the structural mapping. Vandalism is reasonably considered a crime requiring a bare minimum of skill and therefore presents very few barriers to entry. Burglary from vehicle requires perhaps only a small increase in skill above this baseline. Theft and burglary, by contrast, may require a minimum degree of expertise and planning (Wright, Logie, and Decker 1995), though it would be a stretch to describe these as high-skill activities.

Several distinctions also stand out with respect to violent crimes. Notably, several crimes that might be thought of as precursors of violence do not cluster directly with violent crime. For example, criminal threats, violations of court and restraining orders, and threatening phone calls all occupy a cluster along with the catch-all ‘other crime’. Conversely, theft from person (i.e., theft without threat of force) clusters with violent crimes, though in a technical sense it is considered a non-violent crime. Robbery is a small step away from theft from person and one wonders whether routine activities that facilitate the less serious crime naturally lead to the more serious one.

The clustering shown in Figure 7 may also imply something about the ability to generalize crime prevention strategies across crime types. It may be the case that crimes that cluster together in topical space may be successfully targeted with a common set of crime prevention measures. The original premise behind ‘broken windows policing’ was that efforts targeting misdemeanor crimes impacted the likelihood of felony crime because the same people were involved (Wilson and Kelling 1982). It is also possible that policing efforts targeting certain misdemeanor crime types may have an outsized impact on certain felony crime types because they share similar behavioral and situational foundations, whether or not the same people are involved. Figure 7 suggests, for example, that targeting the conditions that support theft from person might impact robberies. Efforts targeting vandalism might impact burglary from vehicle. In general, we hypothesize that the diffusion of crime prevention benefits across crime types should first occur within crime type clusters and only then extend to other crime clusters.

### 6.2. Limitations

There are several limitations to the present study. The first concerns unique constraints on text-based narratives associated with crime event records. These narratives are unlikely to be completely free to vary in a manner similar to other unstructured text systems. Tweets are constrained in terms of the total number of characters allowed. Beyond this physical size constraint, however, there is literally no limit to what can be expressed topically in a Tweet. Additional topical constraints are surely at play in the composition of narrative statements about crime events. For example, the total diversity of crime present in an environment likely has some upper limit (Brantingham 2016). Thus, narratives describing such crimes may also have some topical upper limit. In addition, we should recognize that the narrative text examined here has a unique bureaucratic function. Text-based narratives are presumably aimed at providing justification for the classification of the crime itself. As alluded to above, this likely means that there is a preferred vocabulary that has evolved to provide minimally sufficient justification. Thus we can imagine that there has been a co-evolution of narrative terms and formal crime types that impacts how topics are ultimately resolved. The near complete separation of property from violent crimes in topic space may provide evidence that such is the case.

A second limitation surrounds our ground truth data. We assumed that the official crime type labels applied to crime events are accurate. However, crime type labels may harbor both intentional and unintentional errors (Gove, Hughes, and Geerken 1985; Maltz and Targonski 2002; Nolan, Haas, and Napier 2011). The application of a crime type label is to some extent a discretionary process and therefore the process is open to manipulation. Additionally, benign classification errors both at the time of report taking and data entry are certainly present. If such mislabeling is not accompanied by parallel changes in the event narrative text, then there are sure to be misalignments between official crime types and discovered crime topics. What would be needed is a ground truth crime database curated by hand to ensure that mislabeling of official crime types is kept to a minimum. Curation by hand is not practical in the present case with ~1 million crime records.

The challenge of mislabeling suggests a possible extension of the work presented here. It is conceivable that a pre-trained crime topic model could be used as an autonomous "cross-check" on the quality of official crime type labels. We envision a process whereby a new crime event, consisting of an official crime type label and accompanying narrative text, is fed through the pre-trained topic model. The event is assigned to its most probable topic based on the words occurring in the accompanying narrative text. If there is a mismatch between the officially assigned crime type and the one determined through crime topic assignment, then a alarm might be set for additional review.

More ambitious is the idea that a ground-truth topic model could be used for fully autonomous classification. Here a new event consisting only of narrative text would be evaluated with an official crime type assigned based on the most probable classification from the topic model. No human intervention would be needed. Exploratory work on this process shows, however, that the narrative texts accompanying crime events in our data sample provides too little information for autonomous classification to be accurate at the scale of individual crime types. Police will almost always have more complete information at the time of assigning official crime type labels. While text-based topic models exploit novel information in a novel way, we must conclude for the moment that the crime topic model presented here is insufficient for fully autonomous classification, especially given the legal demands that would be placed on assigned crime types.

Nevertheless, the analyses presented here suggest that larger scale crime classes can be learned automatically from unstructured text descriptions of those crimes. Individual crimes existing as mixtures of different crime topics and, simultaneously, individual crime topics being distributed across nominally different crime types. Reiterating the conceptual connection with traditional topic modeling methods, the situation with crime parallels the idea that a single Tweet may draw on a mixture of different topics, while a single topic may be distributed across many quite distinctive Tweets. Our view is that latent 'crime topics' capture features of the behaviors and situations underlying crimes that are often impractical to observe and almost completely lost when adopting formal crime classifications. Crime topics also hold potential for greater understanding of the situational causes of crime less constrained by the byproducts of formal crime type classifications. Extending causal inferences using crime topics will be the subject of future work.

## 7. Acknowledgements

This work was funded in part by a grant from the US National Science Foundation (DMS-1737770). We thank the Los Angeles Police Department for providing data to support this research. The authors declare no competing interests in this work.

## 9. Tables

Table 1. Examples of official crime classifications and the narrative text tied to the event.

| Official Crime Classification | Accompanying Narrative Text |
|---|---|
| Homicide | VICT IS A [GANG NAME] GANG MEMBER WAS STANDING ON SIDEWALK SPRAY PAINTING GRAFFITI SUSPS DROVE BY THE VICT FIRED SHOTS AT VICT |
| Assault | VICT AND SUBJ ARE MTHR DAUGHTER VICT ATTPT TO DISCIPLINE SUBJ SUBJ BECAME ANGRY AND ATTPT TO CUT VICT |
| Robbery | SUSP ENTERED LOCATION PRODUCED HANDGUN DEMANDED MONEY FROM REGISTER REMOVED PROPERTY FROM LOCATION AND FLED TO UNKNOWN LOCATION |
| Burglary | UNK SUSP ENTERED VICS RESID BY BREAKING SCREEN ON WINDOW WALKED THROUGHTHE RESID EXITED REAR DOOR AND ENTERED DETACHED GARAGE SUSP EXITED WITH PROPERT |
| Burglary-theft from Vehicle | SUSP USING PORCELAIN CHIPS BROKE VEHS WINDOW PRIOR TO SUSP GAINING ENTRY SUSP FLED THE LOC |
| Motor Vehicle Theft | SUSP ENTERED VIC VEH WITH UNK PRY TOOL AND REMOVED PROP FROM VEH SUSP PUNCHED IGNITION SWITCH |
| Theft | S ENTERED CLOTHING STORE AND TOOK APPROX 20 BLUE TSHIRT AND THEN FLED LOCATION WITHOUT PAYING |

Table 2. Counts of events with and without accompanying narrative text by official crime type.

| | No Narrative Text | Narrative Text | Total | Fraction with Narrative Text |
|---|---|---|---|---|
| Robbery | 597 | 53,379 | 53,976 | 0.989 |
| Burglary | 1,320 | 91,260 | 92,580 | 0.986 |
| Homicide | 28 | 1,565 | 1,593 | 0.982 |
| Assault | 1,032 | 45,665 | 46,697 | 0.978 |
| Kidnapping | 45 | 1,707 | 1,752 | 0.974 |
| Grand Theft Person | 230 | 7,754 | 7,984 | 0.971 |
| Theft | 13,326 | 136,117 | 149,443 | 0.911 |
| Burglary-theft from Vehcile | 20,192 | 126,912 | 147,104 | 0.863 |
| Other Miscellaneous Crime | 72,518 | 256,816 | 329,334 | 0.780 |
| Vandalism | 27,630 | 80,038 | 107,668 | 0.743 |
| Arson | 1,111 | 675 | 1,786 | 0.378 |
| Motor Vehicle Theft | 83,521 | 3,730 | 87,251 | 0.043 |

Table 3. *The t*op twenty-five most common words in *the full text corpus consisting of 7,649,164 discrete words*.

| Word | Count | Proportion |
|---|---|---|
| unknown | 635,099 | 0.0830 |
| property | 305,014 | 0.0399 |
| fled | 277,770 | 0.0363 |
| vehicle | 255,609 | 0.0334 |
| location | 202,661 | 0.0265 |
| removed | 197,171 | 0.0258 |
| entered | 143,602 | 0.0188 |
| window | 106,461 | 0.0139 |
| direction | 106,412 | 0.0139 |
| door | 96,918 | 0.0127 |
| residence | 66,576 | 0.0087 |
| front | 57,912 | 0.0076 |
| open | 55,413 | 0.0072 |
| approached | 55,261 | 0.0072 |
| rear | 50,794 | 0.0066 |
| smashed | 45,553 | 0.0060 |
| left | 45,155 | 0.0059 |
| entry | 40,341 | 0.0053 |
| store | 36,515 | 0.0048 |
| stated | 36,068 | 0.0047 |
| object | 35,696 | 0.0047 |
| money | 33,608 | 0.0044 |
| punched | 33,317 | 0.0044 |
| items | 32,354 | 0.0042 |
| face | 31,653 | 0.0041 |

Table 4. Confusion matrix for official crime types by topics. Dominant words in each topic are shown across the top. Row totals reflect the total number of crimes formally classified under each crime type. Column total reflect the total number of crimes clustered within each topic. Boldface numbers are column maxima.

| | 'window' | 'punch' | 'door' | 'card' | 'direct' | 'approach' | 'store' | 'entri' | 'verbal' | 'vehicl' | 'kill' | 'item' | 'damag' | 'lock' | 'check' | 'phone' | 'object' | 'left' | 'properti' | 'resid' | |
|---|---|---|---|---|---|---|---|---|---|---|---|---|---|---|---|---|---|---|---|---|---|
| | 'smash' | 'face' | 'open' | 'credit' | 'unknown' | 'demand' | 'pai' | 'gain' | 'involv' | 'park' | 'state' | 'busi' | 'caus' | 'secur' | 'cash' | 'cell' | 'hard' | 'purs' | 'locat' | 'ransack' | |
| | 'passeng' | 'time' | 'front' | 'info' | 'fled' | 'monei' | 'exit' | 'properti' | 'disput' | 'unlock' | 'threaten' | 'select' | 'paint' | 'cut' | 'forg' | 'call' | 'unknown' | 'return' | 'remov' | 'enter' | |
| | 'side' | 'fist' | 'pri' | 'account' | 'tool' | 'foot' | 'conceal' | 'remov' | 'argument' | 'driver' | 'fear' | 'pai' | 'threw' | 'bike' | 'account' | 'hand' | 'sharp' | 'miss' | 'enter' | 'rear' | |
| | 'rear' | 'struck' | 'pry' | 'permiss' | 'mean' | 'point' | 'merchandis' | 'forc' | 'push' | 'kei' | 'im' | 'conceal' | 'sprai' | 'bicycl' | 'bank' | 'order' | 'scratch' | 'wallet' | 'fled' | 'window' | |
| | 'front' | 'head' | 'rear' | 'obtain' | 'properti' | 'grab' | 'select' | 'access' | 'engag' | 'unknown' | 'call' | 'enter' | 'injuri' | 'tool' | 'monei' | 'violat' | 'break' | 'insid' | 'unknown' | 'bedroom' | |
| | 'driver' | 'kick' | 'unlock' | 'purchas' | 'broke' | 'handgun' | 'enter' | 'unknown' | 'angri' | 'scratch' | 'life' | 'locat' | 'wall' | 'park' | 'busi' | 'ask' | 'smash' | 'discov' | 'unlock' | 'poe' | |
| | 'broke' | 'close' | 'tool' | 'person' | 'remov' | 'gun' | 'walk' | 'made' | 'grab' | 'tire' | 'told' | 'purchas' | 'kick' | 'return' | 'order' | 'grab' | 'type' | 'unattend' | 'poe' | 'screen' | |
| | 'unknown' | 'hit' | 'side' | 'make' | 'pry' | 'fled' | 'attempt' | 'tool' | 'alterc' | 'insid' | 'safeti' | 'bag' | 'visibl' | 'garag' | 'deposit' | 'text' | 'windshield' | 'observ' | 'busi' | 'insid' | |
| | 'remov' | 'multipl' | 'driver' | 'ident' | 'prop' | 'hand' | 'regist' | 'locat' | 'hand' | 'drove' | 'knife' | 'shelf' | 'scratch' | 'miss' | 'attempt' | 'messag' | 'fled' | 'shop' | 'ent' | 'exit' | |
| **Formal Crime Classification** | **T1** | **T2** | **T3** | **T4** | **T5** | **T6** | **T7** | **T8** | **T9** | **T10** | **T11** | **T12** | **T13** | **T14** | **T15** | **T16** | **T17** | **T18** | **T19** | **T20** | **TOTAL** |
| BURGLARY FROM VEHICLE | **29497** | 148 | 6495 | 267 | 4499 | 36 | 22 | **12663** | 7 | **9735** | 25 | 1239 | 243 | **7872** | 81 | 371 | **7547** | 594 | 8138 | 73 | 89552 |
| BURGLARY | 4879 | 51 | **19294** | 128 | 3032 | 77 | 914 | 11604 | 16 | 148 | 89 | 2050 | 246 | 4482 | 308 | 251 | 1704 | 584 | **16463** | **16705** | 83025 |
| BATTERY | 183 | **28972** | 893 | 197 | 1565 | 5147 | 610 | 172 | **26721** | 1427 | 1512 | 241 | 5384 | 415 | 103 | 959 | 517 | 2196 | 717 | 1276 | 79207 |
| PETTY THEFT | 115 | 172 | 1726 | 1872 | **5943** | 1423 | 9969 | 699 | 330 | 1110 | 574 | **4131** | 170 | 5378 | 1258 | **4914** | 102 | **10521** | 14737 | 2259 | 67403 |
| IDENTITY THEFT | 12 | 36 | 242 | **45672** | 195 | 84 | 193 | 1561 | 17 | 156 | 535 | 598 | 43 | 950 | 5087 | 1251 | 5 | 554 | 103 | 104 | 57398 |
| GRAND THEFT | 84 | 138 | 1241 | 1435 | 4946 | 1543 | 1526 | 1310 | 178 | 1173 | 561 | 1666 | 144 | 3773 | 1821 | 1099 | 108 | 5581 | 13289 | 3081 | 44697 |
| ROBBERY | 79 | 4928 | 442 | 77 | 746 | **23112** | 1949 | 139 | 639 | 1231 | 1406 | 669 | 483 | 277 | 451 | 2593 | 577 | 1407 | 2718 | 435 | 44358 |
| VANDALISM ($400 & over) | 5419 | 362 | 1589 | 84 | 2925 | 310 | 168 | 320 | 962 | 7712 | 159 | 321 | **12595** | 788 | 87 | 149 | 6852 | 444 | 998 | 927 | 43171 |
| ASSAULT WITH DEADLY WEAPON | 393 | 7689 | 421 | 82 | 1783 | 8041 | 331 | 74 | 8038 | 5537 | 2655 | 91 | 3136 | 445 | 34 | 157 | 1332 | 1088 | 469 | 883 | 42679 |
| VANDALISM (less than $400) | 6534 | 375 | 1834 | 85 | 2845 | 318 | 142 | 428 | 1184 | 4454 | 210 | 300 | 8318 | 1186 | 83 | 247 | 5423 | 486 | 820 | 1103 | 36375 |
| CRIMINAL THREATS | 53 | 497 | 216 | 64 | 168 | 925 | 131 | 55 | 3434 | 242 | **25035** | 62 | 182 | 68 | 58 | 1234 | 53 | 223 | 61 | 595 | 33356 |
| THEFT FROM VEHICLE - PETTY | 362 | 9 | 1640 | 235 | 2241 | 76 | 24 | 1331 | 17 | 7280 | 27 | 376 | 69 | 478 | 50 | 278 | 91 | 787 | 4980 | 70 | 20421 |
| SHOPLIFTING | 1 | 14 | 72 | 41 | 80 | 48 | **12353** | 10 | 7 | 39 | 6 | 3405 | 9 | 70 | 46 | 143 | 5 | 227 | 1103 | 6 | 17685 |
| THEFT FROM VEHICLE - GRAND | 184 | 3 | 1038 | 68 | 1825 | 36 | 8 | 891 | 6 | 4899 | 13 | 312 | 53 | 452 | 28 | 96 | 77 | 501 | 3625 | 58 | 14173 |
| FORGED OR STOLEN DOCUMENT | 4 | 11 | 8 | 1265 | 53 | 66 | 118 | 57 | 12 | 61 | 51 | 151 | 4 | 7 | **9099** | 26 | 0 | 45 | 65 | 32 | 11135 |
| OTHER MISCELLANEOUS CRIME | 77 | 343 | 303 | 1249 | 211 | 625 | 1280 | 220 | 648 | 1000 | 1350 | 125 | 377 | 376 | 721 | 345 | 45 | 393 | 347 | 487 | 10522 |
| ANNOYING/LEWD/OBSCENE PHONE CALLS/LETTERS | 4 | 730 | 28 | 171 | 35 | 102 | 27 | 13 | 202 | 35 | 3713 | 19 | 301 | 12 | 39 | 3212 | 4 | 107 | 14 | 147 | 8915 |
| VIOLATION OF COURT ORDER | 35 | 324 | 205 | 85 | 85 | 275 | 69 | 38 | 160 | 186 | 1446 | 15 | 56 | 21 | 2592 | 1102 | 5 | 288 | 225 | 1165 | 8377 |
| ATTEMPTED BURGLARY | 771 | 10 | 2174 | 4 | 262 | 21 | 19 | 1861 | 3 | 17 | 14 | 37 | 111 | 274 | 14 | 10 | 213 | 33 | 250 | 945 | 7043 |
| ATTEMPTED ROBBERY | 28 | 749 | 67 | 11 | 69 | 3820 | 172 | 30 | 61 | 166 | 263 | 49 | 79 | 45 | 64 | 304 | 85 | 243 | 206 | 26 | 6537 |
| THEFT FROM PERSON | 17 | 42 | 39 | 61 | 245 | 2979 | 66 | 5 | 125 | 223 | 50 | 45 | 22 | 43 | 37 | 1181 | 9 | 762 | 548 | 28 | 6527 |
| TRESPASSING/LOITERING ON PRIVATE PROPERTY | 129 | 214 | 640 | 294 | 162 | 129 | 107 | 283 | 96 | 96 | 200 | 119 | 92 | 332 | 76 | 23 | 17 | 417 | 1558 | 975 | 5959 |
| VIOLATION OF RESTRAINING ORDER | 42 | 128 | 225 | 57 | 77 | 202 | 38 | 47 | 162 | 161 | 612 | 10 | 32 | 10 | 1007 | 875 | 10 | 111 | 164 | 1268 | 5238 |
| BRANDISHING WEAPON | 25 | 62 | 71 | 3 | 82 | 1249 | 49 | 10 | 761 | 266 | 524 | 8 | 38 | 31 | 3 | 21 | 19 | 70 | 37 | 115 | 3444 |
| CHILD ENDANGERING/NEGLECT | 8 | 195 | 51 | 27 | 15 | 79 | 118 | 19 | 362 | 194 | 905 | 10 | 264 | 19 | 49 | 21 | 10 | 413 | 76 | 402 | 3237 |
| BICYCLE - STOLEN | 6 | 4 | 43 | 12 | 156 | 31 | 31 | 19 | 2 | 11 | 3 | 3 | 3 | 1365 | 1 | 1 | 8 | 151 | 401 | 58 | 2309 |
| EMBEZZLEMENT-GRAND THEFT | 0 | 3 | 0 | 43 | 1 | 18 | 16 | 0 | 2 | 1802 | 16 | 0 | 1 | 0 | 10 | 31 | 0 | 279 | 1 | 1 | 2224 |
| FELONY BATTERY ON POLICE OFFICER | 12 | 822 | 33 | 12 | 9 | 97 | 66 | 26 | 163 | 73 | 95 | 3 | 381 | 6 | 30 | 8 | 10 | 137 | 19 | 5 | 2007 |
| BUNCO - GRAND THEFT | 0 | 5 | 3 | 228 | 33 | 462 | 61 | 10 | 14 | 90 | 142 | 19 | 6 | 7 | 575 | 123 | 0 | 33 | 64 | 44 | 1919 |
| SHOTS FIRED | 29 | 57 | 28 | 13 | 411 | 446 | 9 | 3 | 56 | 325 | 51 | 3 | 91 | 6 | 5 | 0 | 7 | 44 | 117 | 75 | 1776 |
| **TOTAL** | 50313 | 48937 | 42787 | 56078 | 36159 | 56162 | 33729 | 35189 | 45923 | 55982 | 46582 | 17012 | 34995 | 30147 | 25571 | 22355 | 25416 | 30630 | 74287 | 35441 | |

10. Figures

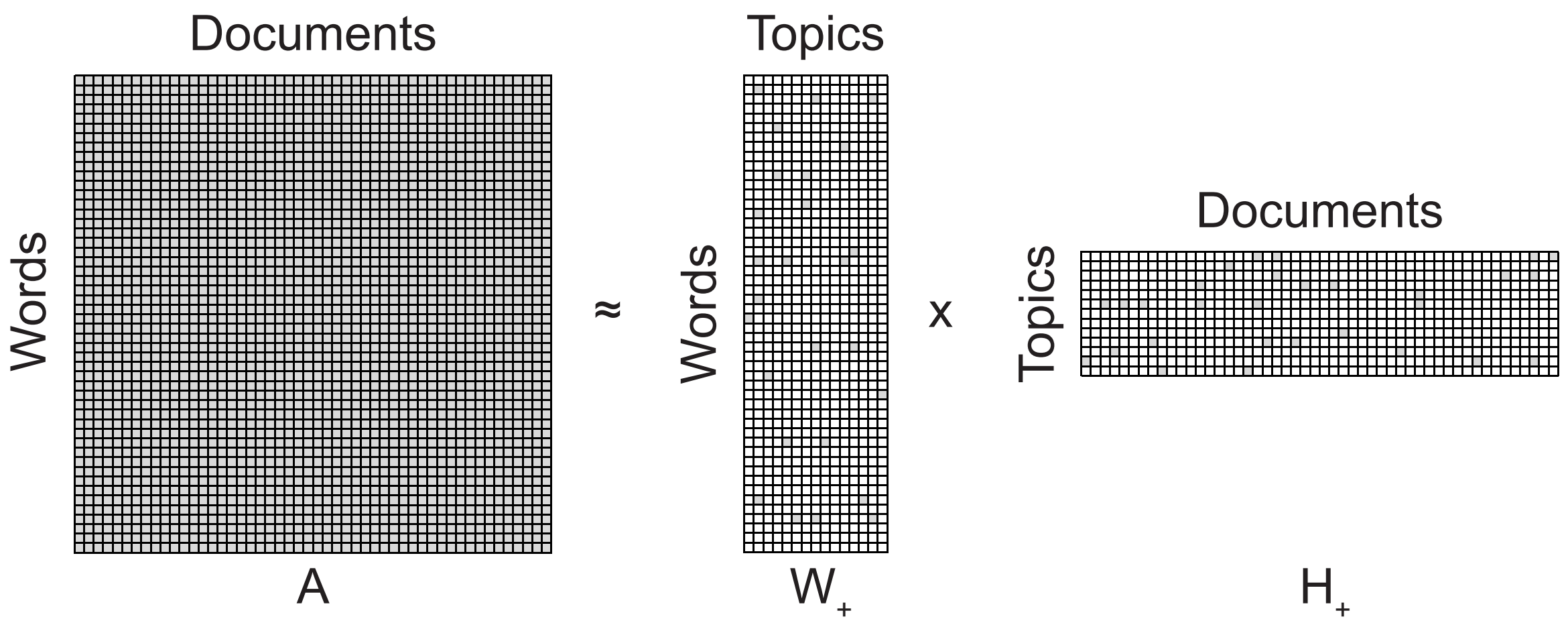

Figure 1. Conceptual illustration of non-negative matrix factorization (NMF) decomposition of a matrix consisting of *m* words in *n* documents into two non-negative matrices of the original *n* words by *k* topics and those same *k* topics by the *m* original documents.

Property Crimes

| A | B | C | D | E | F | G | H | I | J | K | L | M | N | O |
|---|---|---|---|---|---|---|---|---|---|---|---|---|---|---|
| 357,473 | 102,600 | 78,045 | 24,555 | 254,873 | 94,146 | 160,727 | 42,152 | 118,575 | 14,302 | 104,273 | 66,414 | 31,638 | 34,776 | 37,859 |
| (5.7%, 94.3%) | (6.1%, 93.9%) | (6.6%, 93.4%) | (4.4%, 95.6%) | (5.5%, 94.5%) | (5.0%, 95.0%) | (5.8%, 94.2%) | (0.9%, 99.1%) | (7.5%, 92.5%) | (1.9%, 98.1%) | (8.3%, 91.7%) | (7.3%, 92.7%) | (5.9%, 94.1%) | (8.5%, 91.5%) | (10.1%, 89.9%) |
| properti | window | smash | object | properti | door | properti | lock | properti | unlock | properti | properti | remov | locat | direct |
| unknown | smash | window | hard | door | resid | unknown | entri | remov | vehicl | direct | locat | properti | enter | unknown |
| remov | vehicl | vehicl | unknown | remov | open | remov | gain | direct | enter | unknown | remov | vehicl | properti | fled |
| vehicl | object | passeng | scratch | unknown | front | direct | tool | unknown | remov | locat | enter | unsecur | fled | mean |
| window | passeng | properti | vehicl | enter | rear | vehicl | secur | locat | door | remov | fled | unknown | unknown | vehicl |
| fled | unknown | unknown | damag | locat | enter | locat | cut | enter | properti | fled | unknown | permiss | direct | tool |
| direct | side | remov | sharp | fled | pri | fled | vehicl | fled | unknown | enter | vehicl | left | remov | remov |
| enter | hard | side | break | direct | ransack | enter | unknown | vehicl | open | vehicl | unsecur | ent | vehicl | properti |
| door | rear | rear | direct | open | window | lock | remov | unlock | possibl | mean | left | purs | resid | open |
| locat | properti | fled | fled | resid | properti | entri | bike | mean | park | resid | permiss | room | lock | door |

Violent Crimes

| P | Q | R | S | T | U | V | W | X |
|---|---|---|---|---|---|---|---|---|
| 353,726 | 258,395 | 109,207 | 48,141 | 61,066 | 29,593 | 31,473 | 149,188 | 64,409 |
| (61.6%, 38.4%) | (81.7%, 18.3%) | (90.2%, 9.8%) | (90.5%, 9.5%) | (89.9%, 10.1%) | (88.4%, 11.6%) | (91.4%, 8.6%) | (74.8%, 25.2%) | (74.6%, 25.4%) |
| store | punch | verbal | verbal | face | struck | punch | approach | phone |
| pai | verbal | face | disput | punch | fist | face | demand | kill |
| item | face | punch | involv | time | close | time | monei | call |
| punch | approach | involv | argument | fist | caus | kick | state | state |
| approach | involv | disput | engag | struck | head | ground | phone | threaten |
| face | argument | argument | push | head | hit | approach | fear | cell |
| verbal | disput | time | angri | caus | injuri | argument | kill | im |
| exit | time | struck | alterc | close | time | fled | grab | fear |
| locat | push | push | grab | hit | face | began | fled | life |
| time | struck | fist | enrag | injuri | visibl | push | point | text |
| Y | Z | AA | AB | AC | AD | AE | AF | AG |
| 28648 | 35761 | 84779 | 31831 | 52948 | 16090 | 36858 | 32315 | 4543 |
| (54.7%, 45.3%) | (86.9%, 13.1%) | (74.9%, 25.1%) | (73.8%, 26.2%) | (75.6%, 24.4%) | (70.0%, 30.0%) | (78.0%, 22.0%) | (75.2%, 24.8%) | (97.1%, 2.9%) |
| phone | kill | approach | monei | approach | grab | approach | approach | knife |
| cell | state | demand | demand | grab | neck | foot | foot | stab |
| call | threaten | monei | point | foot | purs | fled | fled | pull |
| order | im | grab | handgun | fled | necklac | punch | punch | produc |
| violat | fear | fled | gun | hand | hand | ask | ask | attempt |
| text | call | point | fear | neck | pull | knife | properti | brandish |
| messag | life | foot | gave | purs | chain | state | locat | cut |
| court | told | handgun | approach | pull | arm | properti | hand | pocket |
| time | gonna | gun | properti | properti | fled | locat | state | approach |
| annoi | safeti | properti | give | locat | walk | pocket | push | arm |

Deception-based Property Crime

| AH | AI | AJ | AK | AL |
|---|---|---|---|---|
| 95,331 | 41,328 | 54,003 | 31,289 | 22,714 |
| (8.2%, 91.8%) | (0.7%, 99.3%) | (14.1%, 85.9%) | (11.3%, 88.7%) | (17.8%, 82.2%) |
| store | card | store | item | store |
| pai | credit | pai | pai | merchandis |
| item | info | item | conceal | pai |
| exit | account | conceal | select | exit |
| conceal | purchas | exit | exit | enter |
| select | obtain | select | locat | walk |
| enter | check | merchandis | busi | conceal |
| merchandis | person | enter | enter | properti |
| locat | permiss | locat | regist | select |
| walk | bank | walk | fail | remov |

node
# events
(v%, p%)

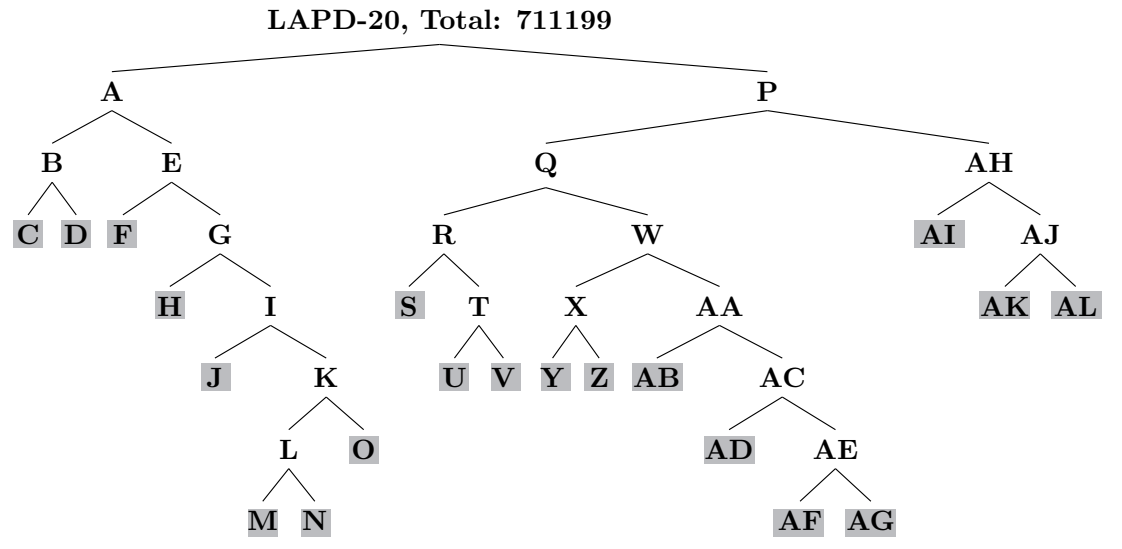

Figure 2. Hierarchical NMF topic structure for the entire corpus of events. The left branch captures property crimes. The right branch captures violent crimes. Deception-based property crimes form a distinct tree in the right branch. Tables show topic labels, number of events in each topic, number of events of the top 40

most frequent crime types in each topic, the percent of events for the topic that are formally classified as violent crime (v%) or property crime (p%), and the top-ten topic words. Terminal leaves of the topic model are marked in gray.

Non-gun Violence

| A | B | C | D | E | F | G | H | I |
|---|---|---|---|---|---|---|---|---|
| 21,875 | 10,600 | 5,960 | 4,640 | 11,275 | 4,114 | 7,161 | 2,197 | 4,964 |
| (1.6%, 98.4%) | (1.9%, 98.1%) | (1.8%, 98.2%) | (2.1%, 97.9%) | (1.2%, 98.8%) | (0.9%, 99.1%) | (1.4%, 98.6%) | (0.3%, 99.7%) | (1.8%, 98.2%) |
| verbal | knife | knife | verbal | head | punch | bottl | bottl | injuri |
| knife | verbal | stab | disput | caus | kick | injuri | glass | caus |
| involv | stab | cut | involv | struck | ground | caus | head | struck |
| argument | involv | attempt | argument | injuri | face | head | threw | head |
| disput | disput | produc | engag | face | time | struck | beer | visibl |
| stab | argument | argument | alterc | punch | approach | glass | hit | hit |
| head | cut | pull | angri | bottl | head | hit | struck | metal |
| caus | engag | kitchen | enrag | hit | began | threw | face | object |
| struck | alterc | arm | hit | kick | fall | visibl | argument | argument |
| face | attempt | hand | struck | glass | fell | argument | verbal | bat |

Gun Violence

| J | K | L | M | N | O | P | Q | R |
|---|---|---|---|---|---|---|---|---|
| 18,333 | 5,730 | 12,603 | 9,367 | 4,201 | 5,166 | 2,208 | 2,958 | 3,236 |
| (5.9%, 94.1%) | (4.9%, 95.1%) | (6.4%, 93.6%) | (7.9%, 92.1%) | (11.8%, 88.2%) | (4.7%, 95.3%) | (1.5%, 98.5%) | (7.0%, 93.0%) | (2.1%, 97.9%) |
| unknown | unknown | vehicl | vehicl | fire | vehicl | vehicl | vehicl | point |
| vehicl | direct | fire | fire | round | drove | drove | exit | gun |
| shot | fled | shot | shot | strike | exit | intention | shot | handgun |
| fire | locat | round | round | shot | shot | hit | fled | state |
| fled | object | strike | strike | unknown | hit | attempt | fire | approach |
| locat | approach | handgun | drove | approxim | fled | collid | passeng | produc |
| direct | shot | gun | unknown | handgun | drive | ram | locat | pull |
| approach | stab | drove | fled | fled | locat | run | stop | kill |
| strike | time | point | exit | approach | stop | att | drive | fled |
| round | foot | exit | locat | multipl | attempt | jump | approach | fear |

node
# events
(h %, a %)

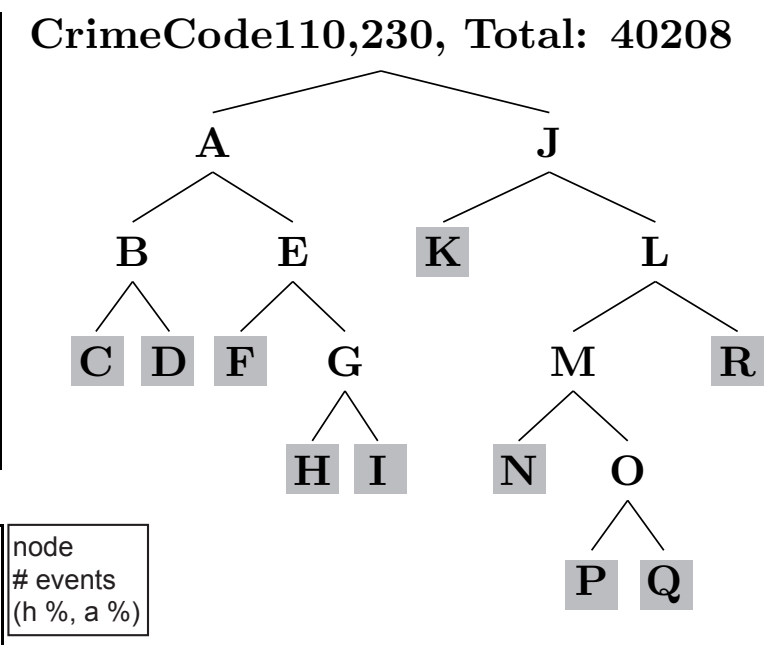

Figure 3. Hierarchical NMF for subset of crimes formally classified as aggravated assault and homicide. Terminal leaves of the topic model are marked in gray.

Non-gun Violence

| A | B | C | D | E | F | G | H | I |
|---|---|---|---|---|---|---|---|---|
| 21,514 | 11,147 | 7,018 | 4,704 | 2,314 | 4,129 | 10,367 | 4,490 | 5,877 |
| verbal | head | bottl | injuri | bottl | punch | knife | verbal | knife |
| knife | caus | injuri | caus | glass | kick | verbal | disput | stab |
| involv | struck | caus | struck | head | ground | stab | involv | cut |
| argument | injuri | head | visibl | threw | face | involv | argument | attempt |
| disput | face | struck | head | beer | time | disput | engag | produc |
| head | punch | glass | metal | struck | head | argument | alterc | argument |
| stab | bottl | hit | hit | hit | approach | cut | angri | pull |
| caus | hit | threw | object | face | began | engag | enrag | kitchen |
| struck | kick | visibl | argument | argument | fall | attempt | hit | arm |
| face | glass | argument | time | verbal | fell | alterc | struck | hand |

Gun Violence

| J | K | L | M | N | O | P | Q | R |
|---|---|---|---|---|---|---|---|---|
| 17,270 | 5,383 | 11,887 | 3,172 | 8,715 | 4,756 | 1,002 | 3,754 | 3,959 |
| unknown | unknown | vehicl | point | vehicl | vehicl | drove | vehicl | fire |
| vehicl | direct | fire | gun | fire | drove | vehicl | exit | round |
| shot | fled | shot | handgun | shot | exit | shot | shot | strike |
| fled | locat | round | state | round | hit | unknown | fled | shot |
| fire | object | strike | approach | strike | shot | strike | hit | unknown |
| locat | approach | handgun | produc | drove | fled | locat | drive | approxim |
| direct | shot | gun | pull | unknown | drive | fire | locat | handgun |
| approach | stab | drove | kill | fled | locat | avoid | intention | fled |
| round | time | point | fled | exit | attempt | jump | attempt | approach |
| strike | struck | exit | fear | locat | stop | hit | stop | gang |

| node |
|---|
| # events |

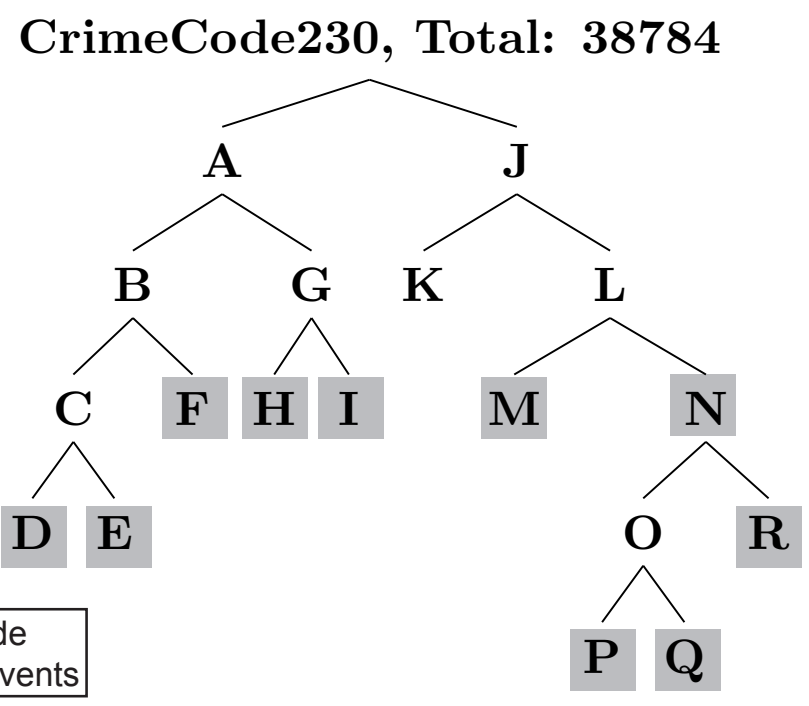

Figure 4. Hierarchical NMF for subset of crimes formally classified as aggravated assaults. Terminal leaves of the topic model are marked in gray.

Gun-Vehicle Homicide

| A | B | C | D | E | F | G | H |
|---|---|---|---|---|---|---|---|
| 996 | 387 | 243 | 48 | 195 | 80 | 115 | 144 |
| vehicl | vehicl | vehicl | seat | vehicl | exit | vehicl | kill |
| fire | kill | exit | passeng | exit | vehicl | drove | gang |
| strike | walk | seat | front | drove | walk | drive | shoot |
| unknown | exit | drive | vehicl | drive | stop | shot | shot |
| multipl | seat | passeng | park | walk | shoot | fire | walk |
| approach | shot | drove | driver | fled | round | struck | car |
| fled | shoot | fire | kill | shot | fire | fled | reason |
| round | drive | walk | open | fire | fled | unknown | member |
| time | gang | park | shot | struck | street | walk | unknown |
| shot | st | front | sat | round | approach | st | confront |

Non-gun Homicide

| R | node<br># events |
|---|---|
| 418 | |
| death | |
| caus | |
| stab | |
| disput | |
| involv | |
| demis | |
| verbal | |
| head | |
| unknown | |
| time | |

Gun Homicide

| I | J | K | L | M | N | O | P | Q |
|---|---|---|---|---|---|---|---|---|
| 609 | 412 | 190 | 102 | 88 | 222 | 197 | 81 | 116 |
| strike | strike | strike | round | strike | fled | di | injuri | di |
| approach | approach | round | numer | multipl | unknown | injuri | succumb | wound |
| multipl | unknown | fire | fire | fire | locat | hospit | scene | hospit |
| fire | fled | numer | strike | time | approach | succumb | multipl | gunshot |
| unknown | round | approach | unknown | approach | foot | wound | time | transport |
| time | fire | multipl | locat | apprchd | time | gunshot | struck | result |
| round | locat | vehicl | approach | gun | direct | transport | pronounc | fire |
| fled | time | stand | fled | stand | multipl | multipl | shot | multipl |
| locat | foot | time | vehicl | head | shot | scene | dead | unknown |
| foot | multipl | unknown | caus | handgun | enter | shot | unknown | shot |

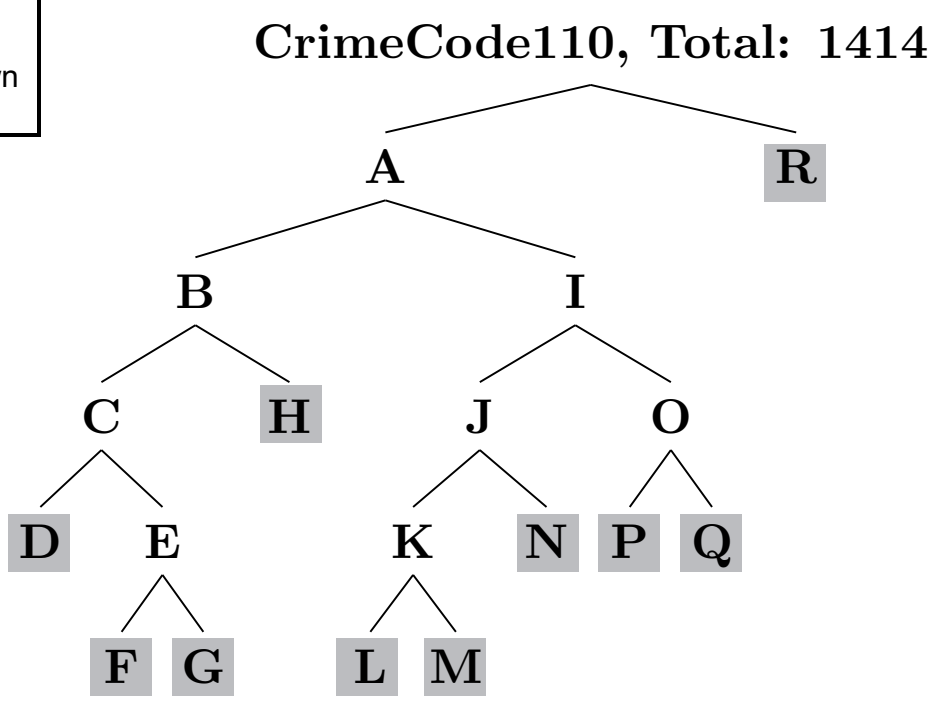

Figure 5. Hierarchical NMF for subset of crimes formally classified as homicides. Terminal leaves of the topic model are marked in gray.

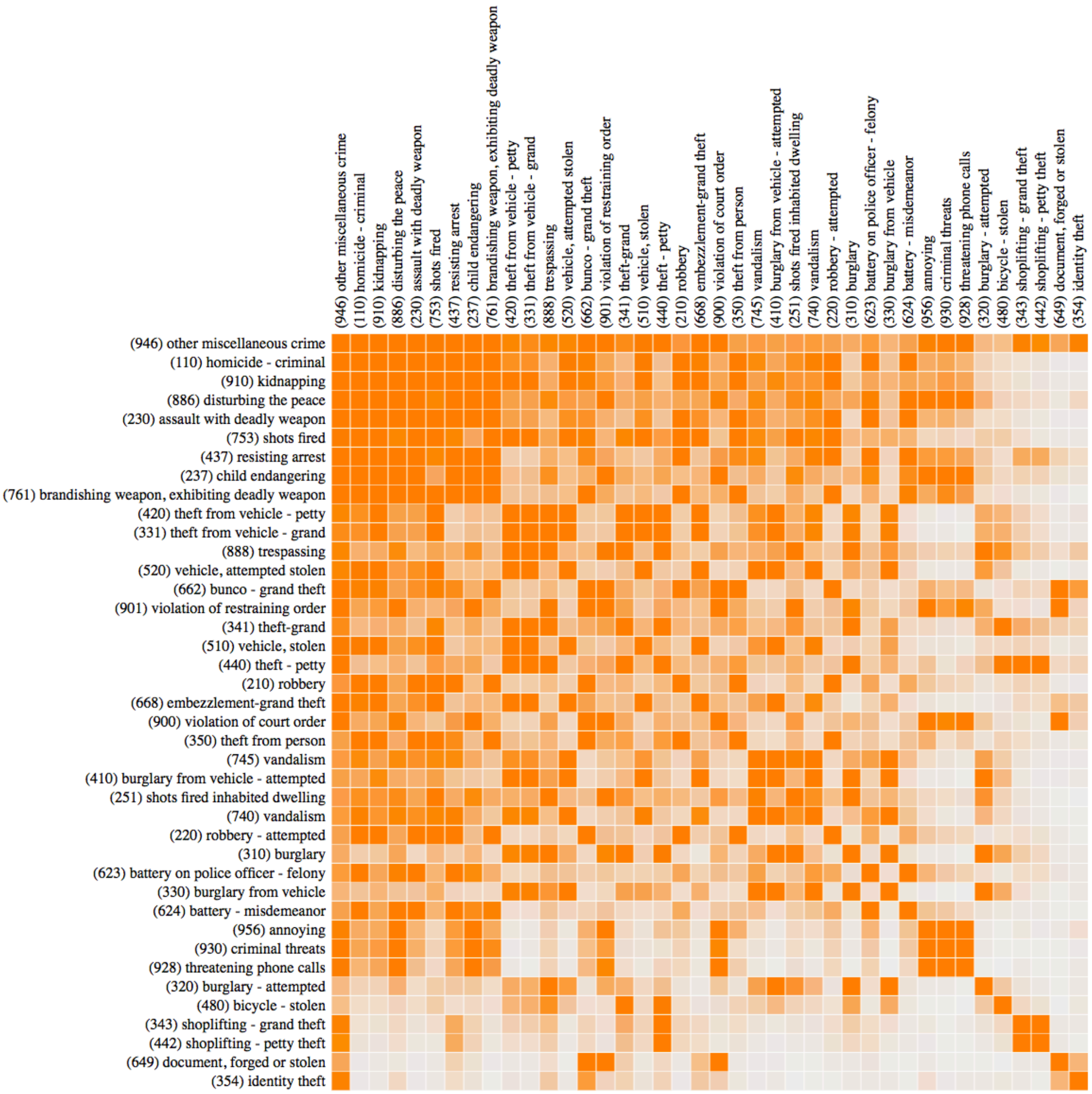

Figure 6. Cosine similarity between crime type pairs sorted in descending order of similarity.

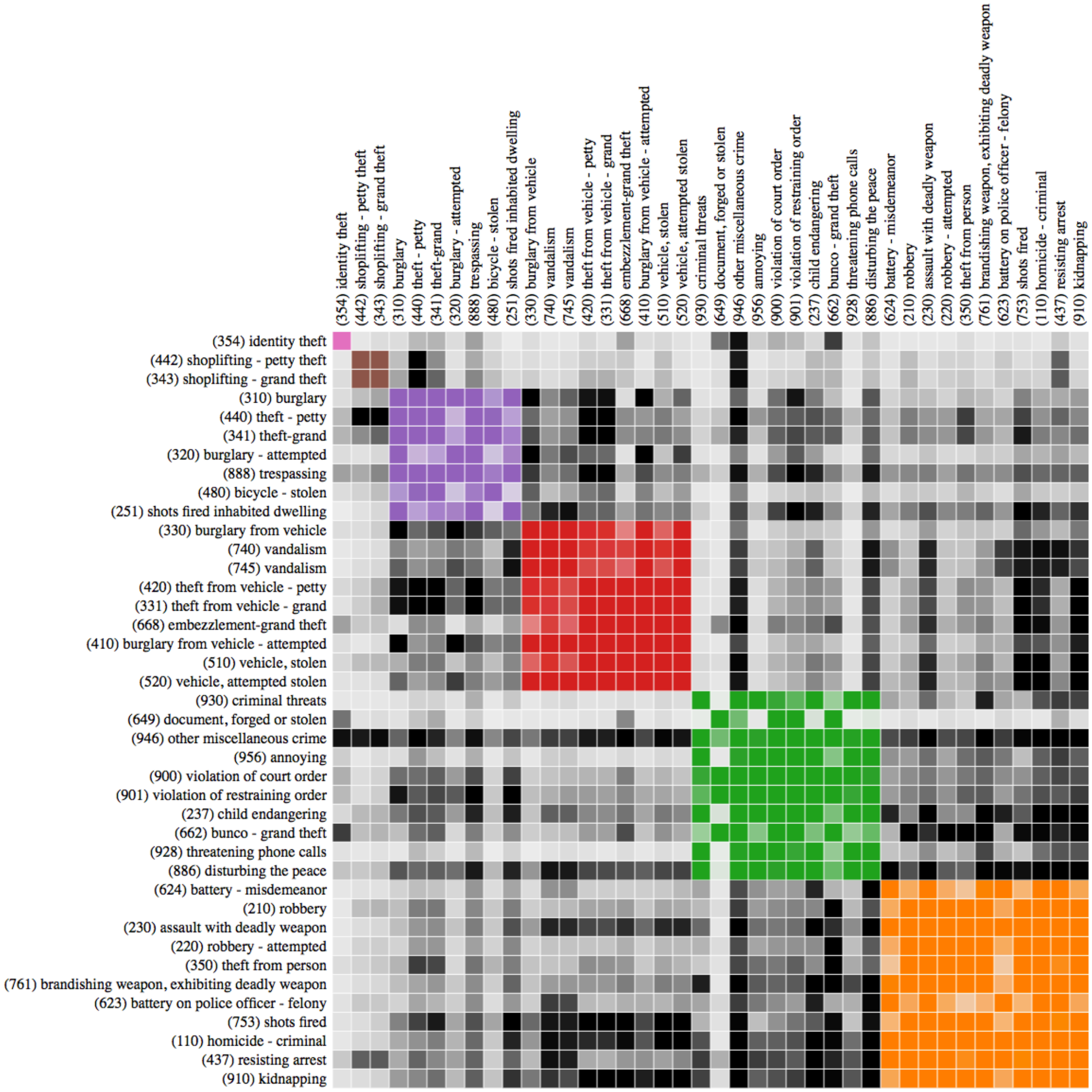

Figure 7. Average linkage clustering for cosine similarity between crime type pairs sorted by cluster proximity.